\documentclass[journal,twoside,web]{ieeecolor}
\usepackage{tmi}
\usepackage{cite}
\usepackage{amsmath,amssymb,amsfonts}
\usepackage{algorithmic}
\usepackage{graphicx}
\usepackage{textcomp}
\usepackage{diagbox}
\usepackage[table]{xcolor} 

\usepackage[utf8]{inputenc} % allow utf-8 input
\usepackage[T1]{fontenc}    % use 8-bit T1 fonts
\usepackage{hyperref}       % hyperlinks
\usepackage{url}            % simple URL typesetting
\usepackage{booktabs}       % professional-quality tables
\usepackage{amsfonts}       % blackboard math symbols
\usepackage{nicefrac}       % compact symbols for 1/2, etc.
\usepackage{microtype}      % microtypography
\usepackage{xcolor}         % colors
\usepackage{amsmath}
\usepackage{threeparttable}
\usepackage{multirow}
\usepackage{graphicx}
\usepackage{tabularx}
\usepackage{amsmath}

\definecolor{ired}{RGB}{255 69 0}
\definecolor{iblue}{rgb}{0.06, 0.75, 1.0}

\definecolor{i2}{rgb}{0.06, 0.75, 1.0}
\definecolor{i1}{rgb}{0.8588, 0.2666, 0.2156}

\def\BibTeX{{\rm B\kern-.05em{\sc i\kern-.025em b}\kern-.08em
    T\kern-.1667em\lower.7ex\hbox{E}\kern-.125emX}}
\begin{document}
\title{Med-R1: Reinforcement Learning for Generalizable Medical Reasoning in Vision-Language Models}
\author{
Yuxiang Lai, Jike Zhong, Ming Li, Shitian Zhao, Yuheng Li, Konstantinos Psounis, \IEEEmembership{Fellow, IEEE}, and Xiaofeng Yang, \IEEEmembership{Member, IEEE}
\thanks{Equal contribution: Yuxiang Lai, Jike Zhong, and Ming Li. Corresponding author: Xiaofeng Yang (email: xiaofeng.yang@emory.edu).}
\thanks{This work is supported in part by the National Institutes of Health under award numbers R01CA272991, R01EB032680, R01DE033512, and U54CA274513.}
\thanks{Yuxiang Lai and Xiaofeng Yang are with the Department of Computer Science and Informatics, Emory University, Atlanta, GA 30322, USA.}
\thanks{Jike Zhong and Konstantinos Psounis are with Department of Computer Science and Department of Electrical and Computer Engineering, University of Southern California, Los Angeles, CA 90089, USA.}
\thanks{Ming Li is with the Department of Computer Science, University of Tokyo, Tokyo 113-8654, Japan.}
\thanks{Shitian Zhao is with the Department of Computer Science, Johns Hopkins University, Baltimore, MD 21218, USA.}
\thanks{Yuheng Li and Xiaofeng Yang are with the Department of Biomedical Engineering, Georgia Institute of Technology and Emory University, Atlanta, GA 30332, USA. Xiaofeng Yang is also with the Department of Radiation Oncology and Winship Cancer Institute, Emory University, Atlanta, GA 30322, USA.}
}
% \author{First A. Author, \IEEEmembership{Fellow, IEEE}, Second B. Author,and Third C. Author, Jr., \IEEEmembership{Member, IEEE}
% \thanks{This paragraph of the first footnote will contain the date on which
% you submitted your paper for review. It will also contain support information,
% including sponsor and financial support acknowledgment. For example, 
% ``This work was supported in part by the U.S. Department of Commerce under Grant BS123456.'' }
% \thanks{The next few paragraphs should contain the authors' current affiliations,
% including current address and e-mail. For example, F. A. Author is with the
% National Institute of Standards and Technology, Boulder, CO 80305 USA (e-mail:author@boulder.nist.gov). }
% \thanks{S. B. Author, Jr., was with Rice University, Houston, TX 77005 USA.
% He is now with the Department of Physics, Colorado State University,
% Fort Collins, CO 80523 USA (e-mail: author@lamar.colostate.edu).}
% \thanks{T. C. Author is with the Electrical Engineering Department,
% University of Colorado, Boulder, CO 80309 USA, on leave from the National
% Research Institute for Metals, Tsukuba, Japan (e-mail: author@nrim.go.jp).}}

\maketitle

\begin{abstract}
% Vision-language models (VLMs) have made significant progress in natural image reasoning, yet their potential in medical imaging remains largely underexplored. Medical vision-language tasks demand accurate image understanding and clinically grounded answers, presenting unique challenges due to the complexity of medical data. Transparency and trustworthiness are critical not only for clinician confidence but also for regulatory approval.
Vision-language models (VLMs) have achieved impressive progress in natural image reasoning, yet their potential in medical imaging remains underexplored. Medical vision-language tasks demand precise understanding and clinically coherent answers, which are difficult to achieve due to complexity of medical data and the scarcity of high-quality expert annotations. These challenges limit the effectiveness of conventional supervised fine-tuning (SFT) and Chain-of-Thought (CoT) strategies that work well in general domains.
To address these challenges, we propose Med-R1, a reinforcement learning (RL)-enhanced VLM designed to improve generalization and reliability in medical reasoning. Med-R1 adopts Group Relative Policy Optimization (GRPO) to encourage reward-guided learning beyond static annotations.
We comprehensively evaluate Med-R1 across \textit{eight} distinct medical imaging modalities.
Med-R1 achieves a \textbf{29.94\% improvement} in average accuracy over its base model Qwen2-VL-2B, and even outperforms Qwen2-VL-72B—a model with \textbf{36$\times$ more parameters}. To assess cross-task generalization, we further evaluate Med-R1 on \textit{five} question types.
% modality recognition, anatomy identification, disease diagnosis, lesion grading, and biological attribute analysis. 
Med-R1 outperforms Qwen2-VL-2B by \textbf{32.06\%} in question-type generalization, also surpassing Qwen2-VL-72B.
We further explore the thinking process in Med-R1, a crucial component of Deepseek-R1. Our results show that omitting intermediate rationales (\textit{No-Thinking Med-R1}) not only improves cross-domain generalization with less training, but also challenges the common assumption that more reasoning always helps. Nevertheless, we also find that the \textit{Think-After Med-R1} variant further improves performance while maintaining interpretability. These findings suggest that in medical VQA, it is not the presence of reasoning itself, but rather its \textit{quality} and \textit{position}.
Together, these results highlight that RL improves medical reasoning and generalization, enabling efficient and reliable VLMs.
\end{abstract}


\section{Introduction}
\label{sec:introduction}
% \IEEEPARstart{T}{he} Vision-language model (VLM) has achieved remarkable progress in reasoning over natural images (e.g., GPT-4o~\cite{hurst2024gpt}, Gemini-1.5~\cite{geminiteam2024gemini15unlockingmultimodal}, and Qwen-VL~\cite{bai2023qwen}), demonstrating impressive capabilities in tasks such as visual question answering (VQA) and multimodal dialogue. These advancements have been driven by large-scale vision-language pretraining and supervised fine-tuning (SFT), which enables models to associate visual content with textual information. However, applying VLMs to medical imaging reasoning remains a significant challenge. Unlike natural images, medical data demands precise interpretation, requiring VLMs to produce not only answer outcomes but also intermediate reasoning paths that are consistent with clinical decision-making logic. For instance, diagnosing a lung nodule in a CT scan necessitates localizing lesions, analyzing morphological features, and integrating patient history—a reasoning chain that must be both accurate and interpretable to gain clinicians' trust. Furthermore, the diversity of medical imaging modalities (e.g., CT, MRI, microscopy) and task types (e.g., diagnosis, lesion grading) poses stringent demands on model generalizability. These challenges raise a critical question: How can we equip VLMs with robust reasoning abilities that generalize across medical domains while maintaining medical reliability?
\IEEEPARstart{V}{ision-language models} (VLMs) have achieved strong performance on natural image understanding tasks such as visual question answering (VQA) and multimodal dialogue~\cite{hurst2024gpt, geminiteam2024gemini15unlockingmultimodal, bai2023qwen}, enabled by large-scale pretraining and supervised fine-tuning (SFT). However, applying VLMs to medical imaging remains challenging due to the need for clinically sound interpretations and decision-making processes. Medical tasks often involve multi-step analysis—e.g., diagnosing a lung nodule may require integrating lesion localization, morphology, and context. In addition, the diversity of imaging modalities (e.g., CT, MRI) and task types (e.g., diagnosis, grading) imposes demands on generalizability. This raises a key question: how can we enable VLMs to perform well across medical domains while ensuring reliable and context-aware behavior?

In this paper, we identify that the limitations of current medical VLMs primarily stem from the inherent drawbacks of Supervised Fine-Tuning (SFT)~\cite{chen2022program,achiam2023gpt}. While SFT has been widely adopted to adapt foundation models to medical imaging~\cite{moor2023med,zhang2023pmc,li2023llava,chen2024huatuogpt,li2025towards}, it suffers from two fundamental issues that hinder medical applicability.
First, SFT inherently biases models toward memorizing task-specific shortcuts rather than learning generalizable reasoning. By directly aligning model outputs with final answers (e.g., diagnostic labels), SFT encourages overfitting to superficial patterns in training data. 
% Second, SFT struggles to cultivate clinically grounded reasoning, even when Chain-of-Thought (CoT) annotations are provided. Medical reasoning demands strict adherence to domain-specific logic (e.g., ruling out differential diagnoses before confirming a malignancy), yet SFT-trained models often generate CoT rationales that merely mimic annotated steps without understanding their clinical significance. This limitation is exacerbated by the prohibitive cost of curating high-quality CoT datasets, which require meticulous annotation by medical experts to capture valid diagnostic workflows.
% Consequently, existing SFT-based medical VLMs~\cite{moor2023med,zhang2023pmc} frequently produce "black-box" predictions—lacking interpretable rationales and failing to adapt to out-of-domain tasks (e.g., from radiology to pathology). These shortcomings pose a major barrier to clinical adoption, where transparency and robustness are non-negotiable.
Second, the scarcity of high-quality Chain-of-Thought (CoT) annotations severely limits the effectiveness of SFT in medical reasoning. Unlike general-domain tasks, where large-scale CoT datasets can be crowdsourced, medical reasoning requires domain-specific logical structuring (e.g., systematically ruling out differential diagnoses before confirming malignancy). However, curating such CoT datasets is prohibitively expensive, as it demands meticulous annotation by experienced medical professionals to ensure diagnostic validity and clinical coherence.
As a result, existing SFT-based medical VLMs~\cite{moor2023med,zhang2023pmc} lack access to high-quality CoT data, leading to shallow reasoning with limited clinical rigor. These models frequently produce "black-box" predictions, struggling to provide traceable reasoning or maintain performance in out-of-domain tasks. This lack of transparency and robustness poses a significant challenge to medical adoption, where explainability and reliability are indispensable requirements.

% To address these challenges, we propose Med-R1, a framework leveraging Reinforcement Learning (RL)~\cite{schulman2017proximal} to enhance both the generalizability and interpretability of medical VLMs. In contrast to SFT, which primarily aligns outputs with fixed supervision, RL promotes exploration of alternative reasoning strategies by leveraging reward-driven feedback, allowing models to construct internal logic without requiring explicit CoT annotations. This approach mitigates overfitting and shortcut learning, as demonstrated by Chu~\etal~\cite{chu2025sft}, who showed that RL-trained vision-language models significantly outperform SFT in cross-domain generalization. Building on recent advancements in RL optimization, we adopt Group Relative Policy Optimization (GRPO)~\cite{shao2024deepseekmath}, a novel strategy that eliminates the need for complex value models required by traditional methods like Proximal Policy Optimization (PPO)~\cite{schulman2017proximal}. GRPO incorporates rule-based rewards derived from medical guidelines (e.g., diagnostic criteria for tumors) and group-relative comparisons to stabilize policy updates. For example, rewards can be designed to prioritize reasoning steps that align with radiological decision trees, ensuring clinically plausible rationales. This dual mechanism not only reduces computational overhead but also enforces adherence to domain-specific constraints, making GRPO uniquely suited for medical reasoning tasks where interpretability and scalability are essential for deployment.

To address these challenges, we propose Med-R1, a reinforcement learning (RL)-based framework for enhancing the generalizability and interpretability of medical VLMs. Unlike SFT, which aligns outputs to fixed supervision and often leads to shortcut learning, RL encourages exploration of diverse reasoning strategies through reward signals—without requiring explicit CoT annotations~\cite{chu2025sft}. We adopt Group Relative Policy Optimization (GRPO)~\cite{shao2024deepseekmath}, a lightweight alternative to PPO (Proximal Policy Optimization)~\cite{schulman2017proximal} that stabilizes training via rule-based rewards and group-relative comparisons. These mechanisms reduce computational overhead while encouraging clinically grounded reasoning, making GRPO well-suited for medical tasks where scalability and reliability are essential.

\begin{figure*}[t]
\vspace{-10pt}
	\centering
\includegraphics[width=1\linewidth]{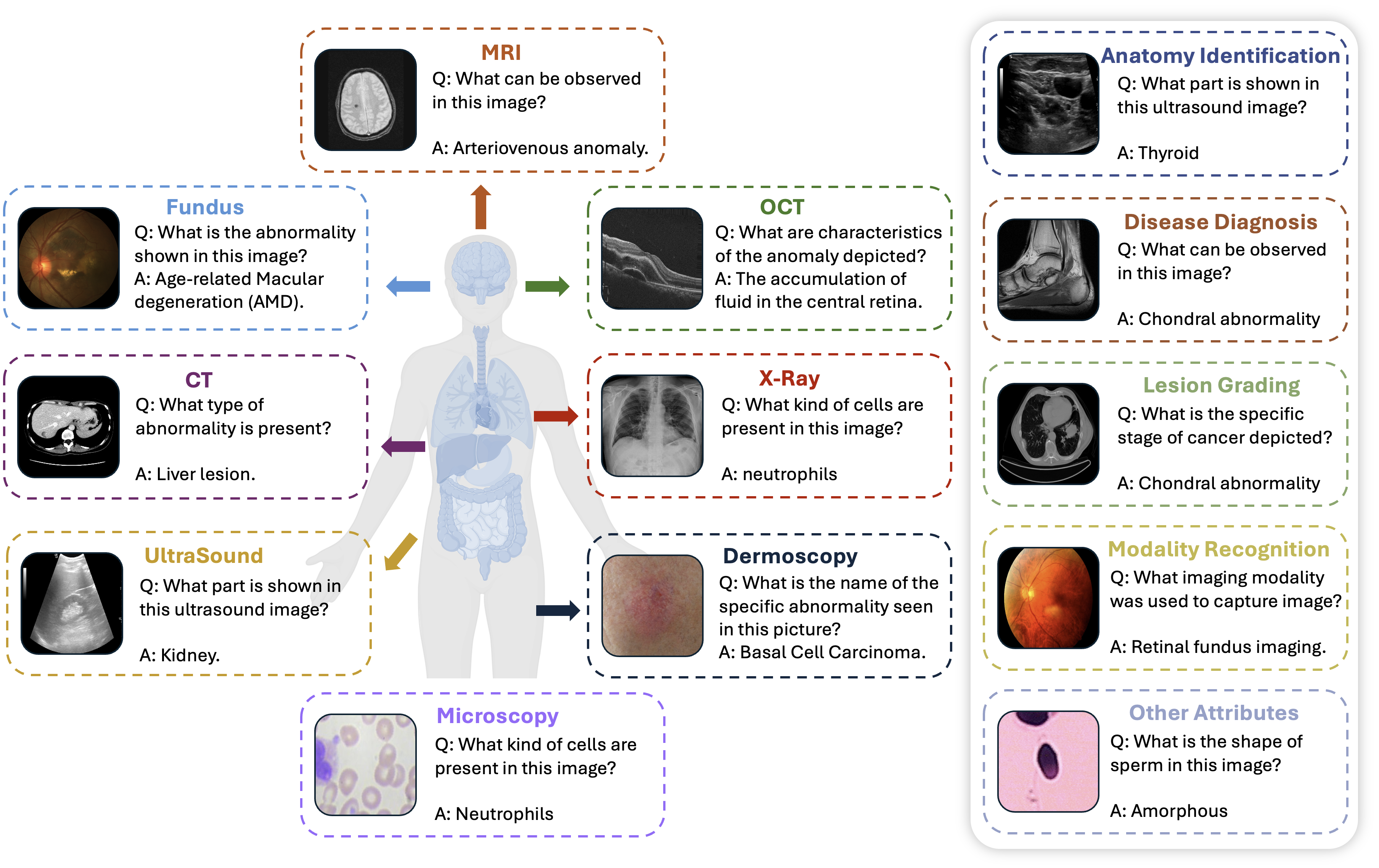}
	\caption{\textbf{ Overview of the Evaluation Framework: Eight Medical Imaging Modalities and Five Medical Vision Question Answering Tasks.} We evaluate Med-R1 across eight distinct medical imaging modalities—Computed Tomography (CT), Magnetic Resonance Imaging (MRI), Ultrasound, Dermoscopy, Fundus Photography, OCT (Optical coherence tomography
), Microscopy Images, and X-ray Imaging—as well as five medical vision question answering tasks: anatomy identification, disease diagnosis, lesion grading, modality recognition, and biological attribute analysis. Example images and corresponding clinical questions illustrate the diversity of medical data and the challenges in developing generalizable vision-language models for automated medical reasoning.}
    
    \label{fig_data}
\end{figure*}

As shown in~\autoref{fig_data}, we evaluate Med-R1 across eight diverse medical imaging modalities, including CT, MRI, Ultrasound, Dermoscopy, Fundus Photography, OCT, Microscopy, and X-ray. These modalities cover a wide range of clinical imaging—from macroscopic anatomy to cellular-level and functional assessments. Med-R1 (2B parameters) achieves 69.91\% average accuracy, a 29.94\% gain over its base model Qwen2-VL-2B, and even outperforms the 72B-parameter Qwen2-VL-72B (\autoref{tab:modality_comparation}), highlighting the benefit of RL-driven adaptation. 
We further assess generalizability across five question types: modality recognition, anatomy identification, disease diagnosis, lesion grading, and biological attribute analysis. Med-R1 improves question-type generalization accuracy by 32.06\% over Qwen2-VL-2B and surpasses Qwen2-VL-72B in this setting as well (\autoref{tab:question_type_comparation}). These demonstrate that RL enhances both parameter efficiency and generalizability in medical VLMs.

% Finally, inspired by recent findings in the thinking process of rule-based RL for image classification~\cite{li2025cls}, we further explore the thinking process in Med-R1. We identify an unexpected behavior in current RL post-training practices. A common assumption in VLM development is that generating step-by-step rationales \textbf{(``Think'')} improves model generalization and performance. However, our findings challenge this view: reasoning strategies learned from general-domain data do not always transfer effectively to clinical tasks. In our experiments, we compare two RL fine-tuning strategies—one where the model is trained to generate intermediate reasoning steps before predicting the answer, and another where the model directly outputs the final answer without any rationales. Counterintuitively, we observe that the latter (\textit{No-Think}) setting consistently outperforms the former across both cross-modality and cross-task generalization benchmarks. These results suggest that effective medical VQA may follow a \textbf{``less is more''} paradigm---where concise, well-aligned predictions outperform verbose but misaligned reasoning. This insight calls for rethinking how reasoning is induced in medical VLMs, especially under domain shift.
Finally, we investigate the impact of intermediate rationales in Med-R1. Conventional wisdom suggests that explicit step-by-step reasoning (``Think'') enhances generalization, yet our results show this is not always true in medical VQA. We compare three RL post-training strategies: (1) \textit{Think}, which generates rationales before answering; (2) \textit{No-Think}, which directly predicts the answer; and (3) our proposed \textit{Think After}, which performs concise reasoning \textit{after} prediction. 
We find that \textit{No-Think} improves generalization across modalities, while \textit{Think} often causes hallucinated rationales due to domain shift. In contrast, \textit{Think After} preserves interpretability without sacrificing accuracy, striking the best balance between reliability and explainability. These findings reveal that in specialized domains, the effectiveness of reasoning depends more on its quality, timing, and domain alignment than its length, challenging the notion that “more thinking is better.”

\begin{figure*}[t]
\vspace{-10pt}
	\centering
\includegraphics[width=1\linewidth]{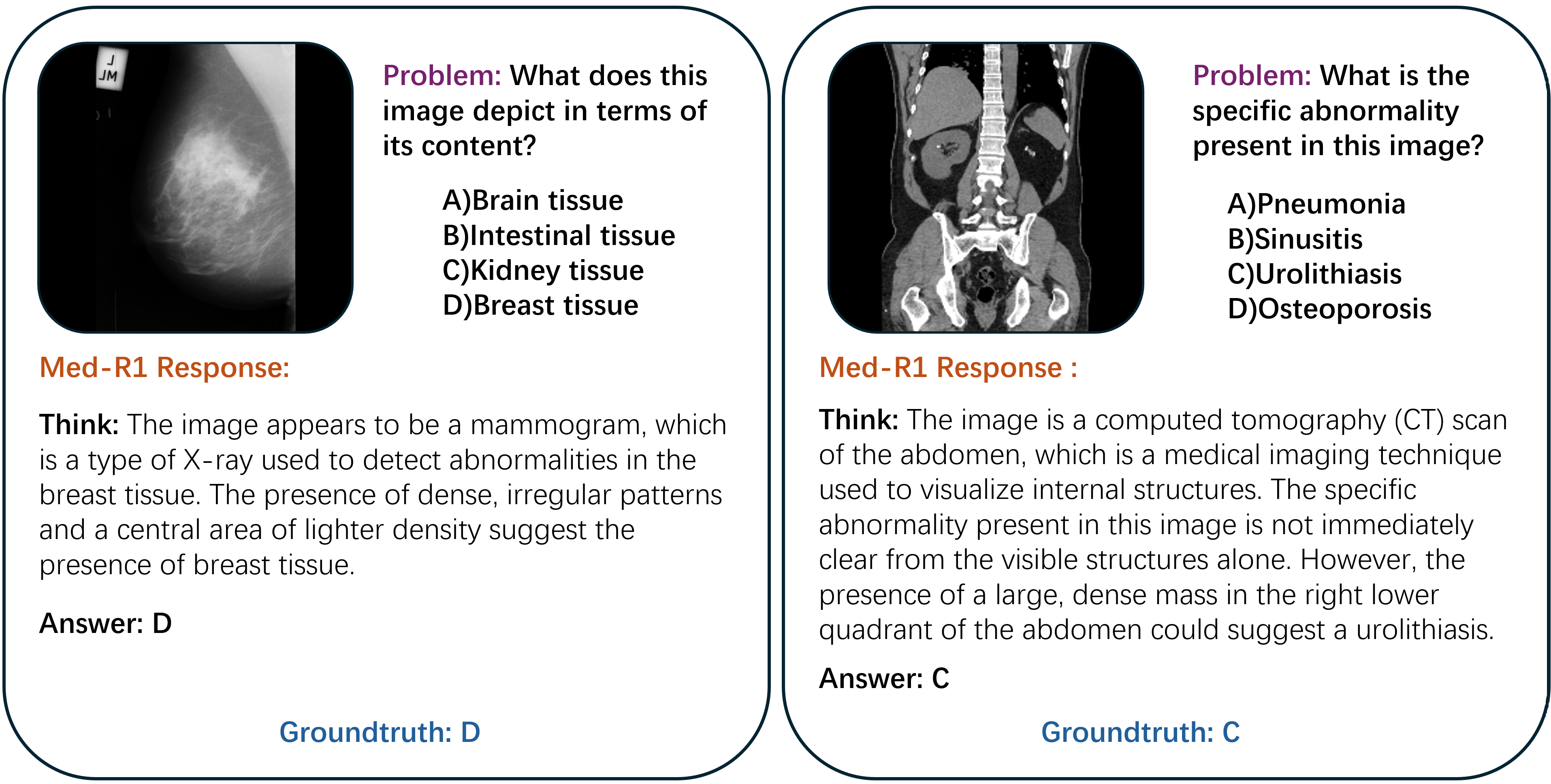}
	\caption{\textbf{Medical VQA examples of Med-R1} The left panel shows a modality recognition task, where the model correctly identifies the presence of breast tissue in a mammogram (X-ray). The right panel illustrates a disease diagnosis task, where Med-R1 accurately detects urolithiasis in an abdominal CT scan. In both cases, the model provides reasoning (“Think”), explaining its decision-making process before selecting the final answer.}
    
    \label{fig_example}
    
\end{figure*}

Med-R1 a comprehensive and systematic study of rule-based RL
for medical reasoning across eight imaging modalities (\autoref{fig_example}). We summarized our contribution as follows:

\begin{enumerate}
 \item \textbf{Multi-modality and multi-task medical reasoning VLM.} We propose \textit{Med-R1}, a comprehensive and systematic study of rule-based reinforcement learning for medical reasoning, supporting \textbf{eight imaging modalities} (CT, MRI, Ultrasound, etc.) across \textbf{five distinct clinical tasks}. We demonstrate that RL-based fine-tuning effectively promotes modality-specific as well as cross-modality reasoning in the medical domain without the need for token-level supervision as in SFT. The proposed Med-R1 is capable of generating step-by-step, accurate, and plausible explanations.

  % \item \textbf{Generalizable Medical VLM with Multi-Modal Explicit Reasoning:} 
  % We propose Med-R1, the first vision-language model supporting \textbf{eight medical imaging modalities} (CT, MRI, Ultrasound, etc.), capable of generating step-by-step and clinically plausible explanations without task-specific retraining. Unlike prior SFT-based models limited to single modalities and black-box answers, Med-R1 achieves cross-domain generalization through RL-driven exploration.

  % \item \textbf{Emergent Reasoning via Clinically Constrained RL:} 
  % We adopt \textbf{Group Relative Policy Optimization}, a RL strategy that eliminates the need for explicit reasoning supervision. GRPO integrates \textit{rule-based rewards} (e.g., enforcing adherence to diagnostic guidelines) and \textit{group-relative comparisons}, enabling models to learn clinically plausible reasoning paths from final-answer labels alone.

  \item \textbf{Robust Generalization with Efficiency.} We show that alongside the solid modality and task-specific performance, Med-R1 exhibits strong \textbf{generalization}. Med-R1 outperforms the base model by \textbf{29.94\%} and SFT baselines by \textbf{15.84\%} in average generalization accuracy across modalities. In cross-task settings, it outperforms the base model and SFT baseline by \textbf{32.06\%} and \textbf{11.25\%} respectively. Moreover, Med-R1 surpasses other larger generic or medical-specific models including Qwen2-VL-72B, and MedVInT-7B, warranting its efficiency and reliability for real-world deployment.

  % \item \textbf{Challenging the ``Think is Better'' Assumption:} 
  %   Through systematic evaluation, we show that \textbf{explicit rationale generation (``Think'')}---a common strategy in general-domain VLMs---can impair performance when transferred to medical tasks. This empirical finding supports a \textbf{``less is more''} paradigm, highlighting the need to rethink the unsupervised reasoning RL in domain-specific VQA.
\item \textbf{Rethinking the ``More Thinking is Better'' Assumption:}  
Our results challenge the common belief that generating longer or more explicit reasoning chains necessarily improves generalization. We find that reinforcement learning without explicit reasoning often yields higher accuracy, as free-form reasoning learned from general-domain data can induce hallucinations under domain shift. However, this may reduce reliability and interpretability in medical applications. To address this, we introduce \textit{Think-After}—a reasoning scheme where the model provides rationalization for its chosen answer \textit{after} the prediction. This design preserves interpretability while mitigating the instability introduced by lengthy reasoning chains, offering a balance between accuracy and explainability. Our findings suggest that the \textit{quality} and \textit{timing} of reasoning are critical for robust generalization.

\end{enumerate}

\section{Related Works}
\smallskip\noindent\textbf{General VLMs and Medical VLMs.}
General-purpose VLMs such as CLIP~\cite{radford2021learning} and BLIP-2~\cite{li2023blip} have advanced natural image-text understanding via large-scale pretraining, but struggle with domain-specific tasks like medical reasoning. Recent efforts adapt VLMs through supervised fine-tuning (SFT) on medical datasets, as seen in LLaVA-Med~\cite{li2023llava} and Med-Flamingo~\cite{moor2023med}. While effective in-domain, these models often overfit to narrow corpora and lack generalization across modalities or task types. Our work addresses this limitation by introducing RL for scalable, modality-agnostic adaptation.

\smallskip\noindent\textbf{Reinforcement Learning for Post-Training.}
RL has shown promise for aligning language models with desired behavior via reward feedback~\cite{ouyang2022training, bai2022constitutional}. In vision-language tasks, RL improves VQA accuracy~\cite{chu2025sft} and reduces hallucination~\cite{shao2024deepseekmath}, but often relies on complex reward models or costly human supervision. GRPO~\cite{shao2024deepseekmath} offers a scalable alternative by using rule-based rewards and group-relative comparisons. We extend this approach to medical VQA, enabling efficient adaptation without modality-specific supervision.

\smallskip\noindent\textbf{Medical Reasoning and Interpretability}  
Interpretable reasoning is critical in medical AI, with recent work exploring CoT prompting~\cite{wei2022chain} and program-guided logic~\cite{chen2022program}. However, such methods rely on costly expert annotations~\cite{li2024abdomenatlas}, limiting scalability in clinical domains. Reinforcement learning offers an alternative by enabling emergent reasoning without explicit supervision. Concurrently, MedVLM-R1~\cite{pan2025medvlm} also applies GRPO-based RL to radiology VQA, which is focuses on a single radiology-specific setting (training on
MRI and testing on CT and X-ray) and reports generalization within radiology, using roughly 600 training
samples. Nonetheless, it highlights the growing interest in RL for medical VLMs.

\section{Method}

% Our method primarily involves tuning a pre-trained large VLM base model with Group Robust Preference Optimization (GRPO) \citet{shao2024deepseekmathpushinglimitsmathematical}, a popular RL-based post-training algorithm.
We adopt rule-based RL to encourage multimodal reasoning and generalization in medical domains. Recent studies have demonstrated that RL can incentivize emergent logical reasoning and generalization in multimodal tasks such as mathematical reasoning \cite{huang2025visionr1incentivizingreasoningcapability} and visual navigation \cite{chu2025sft}. Building on these insights, we extend its application to the medical domain and systematically assess its effectiveness in this context. Specifically, we leverage the popular RL-based post-training method GRPO \cite{shao2024deepseekmathpushinglimitsmathematical} to train a large base MLLM \cite{bai2025qwen25vltechnicalreport} across 8 different modalities for medical reasoning and compare it with zero-shot and SFT performance of popular existing VLMs. We introduce the details of the SFT and RL algorithm, our reward design, and data structure below.

\smallskip\noindent\textbf{Supervised Fine-tuning:} We performed supervised fine-tuning (SFT) on the Qwen2-VL-2B and Qwen2.5-VL-3 models. 
Specifically, we fine-tuned the model using mixed-precision training (bfloat16) with gradient accumulation and checkpointing enabled for memory efficiency. The maximum sequence length was set to 4096 tokens, allowing the model to process full visual-text reasoning sequences. Training was conducted for one epoch with a learning rate of $2\times10^{-5}$, cosine learning-rate scheduling, and a warm-up ratio of 0.1. The per-device batch size was 4, and the gradient accumulation step was 16, resulting in an effective batch size of 64.

\subsection{Group Relative Policy Optimization (GRPO)}
\smallskip\noindent\textbf{Overview:} RL-based algorithms such as PPO \cite{schulman2017proximal} and GRPO \cite{shao2024deepseekmath} belong to a family of fine-tuning and alignment strategies explicitly designed to enhance models' reasoning capacities. Unlike supervised fine-tuning, which directly optimizes maximum likelihood, these RL-based methods instead optimize the policy gradient using reward signals, encouraging reasoning by exploring a much larger solution space. GRPO is closely related to PPO but differs in two key aspects: first, GRPO estimates the advantage using group-based estimation rather than a value function; second, it employs a set of fixed rules as the reward signal instead of a learned reward model. These optimizations make GRPO 50\% more resource- and computation-efficient than PPO \cite{shao2024deepseekmath}.

\smallskip\noindent\textbf{Definition:} 
Formally, let \( P(Q) \) denote the question set used for training, where \( q \) is a sampled question in the current iteration. 
Let \( \pi_{\theta_{\text{old}}} \) and \( \pi_{\theta_{\text{new}}} \) denote the old policy and current (new) policy, respectively, where \( o \) is a complete response sampled from a policy. 
Let \( \pi_{\theta_{\text{ref}}} \) denote the reference policy, which in practice is the frozen base MLLM. 
Let \( G \) be the number of responses sampled per question in each iteration. The GRPO objective is given by:
% \begin{align*}
% &\mathcal{J}_{\text{GRPO}}(\theta) = 
% \mathbb{E}_{q \sim P(Q), \{o_i\}_{i=1}^{G} \sim \pi_{\theta_{\text{old}}}(O | q)}  \\
% & \frac{1}{G} \sum_{i=1}^{G} \Bigg( \min \Bigg( \frac{\pi_{\theta_{\text{new}}}(o_i | q)}{\pi_{\theta_{\text{old}}}(o_i | q)} A_i, \\
% &\text{clip} \Bigg( \frac{\pi_{\theta_{\text{new}}}(o_i | q)}{\pi_{\theta_{\text{old}}}(o_i | q)}, 1 - \epsilon, 1 + \epsilon \Bigg) A_i \Bigg) \notag - \beta \mathbb{D}_{KL} \Big( \pi_{\theta_{\text{new}}} \Big\| \pi_{\text{ref}} \Big) \Bigg),
% \tag{1}
% \end{align*}
\begin{align}
\mathcal{J}_{\text{GRPO}}(\theta) = 
&\ \mathbb{E}_{q \sim P(Q), \{o_i\}_{i=1}^{G} \sim \pi_{\theta_{\text{old}}}(O \mid q)}  \notag \\
&\quad \frac{1}{G} \sum_{i=1}^{G} \Bigg[ 
    \min \Bigg( 
        \frac{\pi_{\theta_{\text{new}}}(o_i \mid q)}{\pi_{\theta_{\text{old}}}(o_i \mid q)} A_i, \notag \\
&\qquad \text{clip} \left( 
        \frac{\pi_{\theta_{\text{new}}}(o_i \mid q)}{\pi_{\theta_{\text{old}}}(o_i \mid q)}, 
        1 - \epsilon, 1 + \epsilon 
    \right) A_i 
    \Bigg) \notag \\
&\quad\quad - \beta\, \mathbb{D}_{\text{KL}} \Big( \pi_{\theta_{\text{new}}} \Big\| \pi_{\text{ref}} \Big) 
\Bigg] \label{eq:grpo}
\end{align}
% where \(\dfrac{\pi_\theta(o_i \mid q)}{\pi_{\theta_{\text{old}}}(o_i \mid q)}\) is the 
where \( \frac{\pi_{\theta}(o_i | q)}{\pi_{\theta_{\text{old}}}(o_i | q)} \) is the
policy ratio, and \( A_i \) is the estimated advantage, and $\epsilon$ is the clipping threshold for policy updates
The KL divergence term \cite{Kullback1951KL} regularizes the policy update, ensuring that \( \pi_{\theta} \) does not deviate excessively from the reference model \( \pi_{\theta_{\text{ref}}} \). Unlike PPO, which uses a critic model to estimate the advantage \( A_i \) for a single response \( o \), GRPO estimates the relative advantage by sampling a group of responses \( \{o_i\}_{i=1}^G \) and normalizing their rewards within the group to compute a relative advantage \cite{shao2024deepseekmath, guo2025deepseek}. Each reward is calculated based on rules without reward models. We detail reward design below.

\begin{table*}[h]
    \vspace{-20pt}
    \centering
    \caption{\textbf{Cross-Modality Generalization Performance of Med-R1 with RL Post-Training.} Accuracy (\%) across eight medical imaging modalities, where rows indicate training modalities and columns test modalities. Darker cell shades indicate higher accuracy for corresponding training-test pairs in each column. The base model is Qwen2.5-VL-3B.
    % The "Overall" column/row reports average performance per training and test domain.
    }

    \setlength{\tabcolsep}{2.5pt}
    \begin{tabular}{p{0.09\linewidth} | *{8}{>{\centering\arraybackslash}p{0.09\linewidth}} | >{\centering\arraybackslash}p{0.09\linewidth} }
    \toprule
    \multicolumn{10}{c}{\textbf{RL fine-tuned Qwen2.5-VL-3B}} \\
    \midrule
    \diagbox{Train}{Test} & CT & MRI & X-Ray & Ultrasound & Dermoscopy & Fundus & OCT & Microscopy & Overall \\
    \midrule
    \textbf{CT} & \cellcolor{ired!80}$94.35\pm0.79$ & \cellcolor{ired!60}$70.85\pm1.10$ & \cellcolor{ired!80}$86.50\pm1.70$ & $35.44\pm2.05$ & \cellcolor{ired!40}$66.77\pm2.53$ & \cellcolor{ired!60}$76.41\pm2.51$ & \cellcolor{ired!80}$84.08\pm2.48$ & \cellcolor{ired!40}$66.58\pm2.79$ & $72.62\pm0.65$ \\
    \textbf{MRI} & \cellcolor{ired!60}$72.79\pm1.53$ & \cellcolor{ired!100}$98.57\pm0.29$ & \cellcolor{ired!80}$83.72\pm1.80$ & $38.38\pm2.07$ & \cellcolor{ired!40}$68.30\pm2.53$ & \cellcolor{ired!60}$79.78\pm2.32$ & \cellcolor{ired!80}$85.97\pm2.36$ & \cellcolor{ired!40}$67.12\pm2.79$ & $74.33\pm0.64$ \\
    \textbf{X-Ray} & \cellcolor{ired!60}$78.90\pm1.39$ & \cellcolor{ired!20}$59.73\pm1.20$ & \cellcolor{ired!80}$93.50\pm1.21$ & $38.28\pm2.10$ & \cellcolor{ired!40}$65.85\pm2.57$ & \cellcolor{ired!60}$76.05\pm2.55$ & \cellcolor{ired!60}$78.89\pm2.77$ & \cellcolor{ired!40}$65.14\pm2.79$ & $69.54\pm0.68$ \\
    \textbf{Ultrasound} & \cellcolor{ired!40}$62.20\pm1.67$ & \cellcolor{ired!40}$63.00\pm1.19$ & \cellcolor{ired!60}$79.57\pm1.98$ & \cellcolor{ired!100}$98.84\pm0.46$ & \cellcolor{ired!40}$64.09\pm2.57$ & \cellcolor{ired!60}$72.22\pm2.60$ & \cellcolor{ired!60}$78.54\pm2.77$ & \cellcolor{ired!40}$69.82\pm2.75$ & $73.53\pm0.65$ \\
    \textbf{Dermoscopy} & \cellcolor{ired!40}$64.30\pm1.65$ & \cellcolor{ired!40}$60.39\pm1.18$ & \cellcolor{ired!80}$80.56\pm1.92$ & $40.21\pm2.12$ & \cellcolor{ired!80}$84.99\pm1.91$ & \cellcolor{ired!60}$71.86\pm2.69$ & \cellcolor{ired!60}$74.76\pm2.89$ & \cellcolor{ired!40}$65.23\pm2.75$ & $67.79\pm0.70$ \\
    \textbf{Fundus} & \cellcolor{ired!40}$68.59\pm1.59$ & \cellcolor{ired!40}$61.32\pm1.19$ & \cellcolor{ired!80}$81.98\pm1.86$ & $37.42\pm2.10$ & \cellcolor{ired!40}$67.84\pm2.49$ & \cellcolor{ired!80}$90.62\pm1.68$ & \cellcolor{ired!60}$79.13\pm2.71$ & \cellcolor{ired!40}$66.76\pm2.79$ & $69.21\pm0.67$ \\
    \textbf{OCT} & \cellcolor{ired!80}$81.06\pm1.34$ & \cellcolor{ired!60}$73.92\pm1.06$ & \cellcolor{ired!80}$83.65\pm1.80$ & $35.78\pm2.03$ & \cellcolor{ired!40}$68.99\pm2.53$ & \cellcolor{ired!80}$80.87\pm2.32$ & \cellcolor{ired!100}$98.70\pm0.77$ & \cellcolor{ired!40}$67.12\pm2.79$ & $73.76\pm0.65$ \\
    \textbf{Microscopy} & \cellcolor{ired!40}$65.50\pm1.62$ & \cellcolor{ired!40}$62.34\pm1.21$ & \cellcolor{ired!80}$80.31\pm1.95$ & $37.37\pm2.10$ & \cellcolor{ired!40}$64.24\pm2.60$ & \cellcolor{ired!60}$71.40\pm2.64$ & \cellcolor{ired!60}$77.36\pm2.83$ & \cellcolor{ired!80}$88.02\pm1.94$ & $68.32\pm0.69$ \\
    \midrule
    \textbf{Overall} & $73.46\pm1.51$ & $68.77\pm1.12$ & $83.72\pm1.79$ & $45.21\pm2.14$ & $68.88\pm2.45$ & $77.40\pm2.50$ & $82.18\pm2.53$ & $69.47\pm2.74$ & $71.14\pm0.67$ \\
    \bottomrule
\end{tabular}
    \begin{tablenotes}
        \item CT - Computed Tomography; MRI -  Magnetic Resonance Imaging; US - Ultrasound; Der - Dermoscopy; FP - Fundus Photography.
        \item OCT - Optical Coherence Tomography; Micro - Microscopy Images; X-Ray - X-Ray Imaging
    \end{tablenotes}
    \label{tab:modality_generalize}
\end{table*}

\begin{table*}[t]
    \vspace{-10pt}
    \centering
    \caption{\textbf{Cross-Modality Generalization Performance of Med-R1 with No-think RL Post-Training.} Accuracy (\%) across eight medical imaging modalities, where rows indicate training modalities and columns test modalities. Darker cell shades indicate higher accuracy for corresponding training-test pairs in each column. The base model is Qwen2.5-VL-3B}
    \setlength{\tabcolsep}{2.5pt}
    \begin{tabular}{p{0.09\linewidth} | *{8}{>{\centering\arraybackslash}p{0.09\linewidth}} | >{\centering\arraybackslash}p{0.09\linewidth} }
    \toprule
    \multicolumn{10}{c}{\textbf{No-Think RL fine-tuned Qwen2.5-VL-3B}} \\
    \midrule
    \diagbox{Train}{Test} & CT & MRI & X-Ray & Ultrasound & Dermoscopy & Fundus & OCT & Microscopy & Overall \\
    \midrule
    \textbf{CT} & \cellcolor{ired!100}$98.27\pm0.45$ & \cellcolor{ired!60}$74.77\pm1.04$ & \cellcolor{ired!80}$85.33\pm1.70$ & $37.61\pm2.05$ & \cellcolor{ired!40}$69.07\pm2.53$ & \cellcolor{ired!60}$78.87\pm2.46$ & \cellcolor{ired!80}$87.26\pm2.30$ & \cellcolor{ired!40}$67.75\pm2.75$ & $74.87\pm0.64$ \\
    \textbf{MRI} & \cellcolor{ired!40}$69.70\pm1.56$ & \cellcolor{ired!100}$99.61\pm0.15$ & \cellcolor{ired!80}$81.73\pm1.89$ & $34.91\pm2.03$ & \cellcolor{ired!60}$75.19\pm2.34$ & \cellcolor{ired!80}$80.51\pm2.32$ & \cellcolor{ired!80}$89.98\pm2.00$ & \cellcolor{ired!40}$67.48\pm2.75$ & $74.89\pm0.65$ \\
    \textbf{X-Ray} & \cellcolor{ired!80}$80.10\pm1.36$ & \cellcolor{ired!40}$69.78\pm1.14$ & \cellcolor{ired!100}$95.79\pm0.99$ & $34.33\pm2.05$ & \cellcolor{ired!60}$71.29\pm2.41$ & \cellcolor{ired!80}$81.79\pm2.28$ & \cellcolor{ired!80}$83.14\pm2.54$ & \cellcolor{ired!40}$68.65\pm2.79$ & $73.11\pm0.63$ \\
    \textbf{Ultrasound} & \cellcolor{ired!20}$56.40\pm1.70$ & \cellcolor{ired!40}$64.93\pm1.18$ & \cellcolor{ired!80}$80.25\pm1.92$ & \cellcolor{ired!100}$100.0\pm0.00$ & \cellcolor{ired!40}$68.38\pm2.53$ & \cellcolor{ired!60}$75.32\pm2.60$ & \cellcolor{ired!60}$76.30\pm2.89$ & \cellcolor{ired!60}$71.71\pm2.66$ & $74.16\pm0.66$ \\
    \textbf{Dermoscopy} & \cellcolor{ired!20}$55.45\pm1.70$ & \cellcolor{ired!40}$66.48\pm1.15$ & \cellcolor{ired!80}$80.37\pm1.95$ & $37.42\pm2.12$ & \cellcolor{ired!80}$92.88\pm1.38$ & \cellcolor{ired!60}$71.86\pm2.69$ & \cellcolor{ired!60}$76.53\pm2.83$ & \cellcolor{ired!40}$67.12\pm2.70$ & $68.51\pm0.69$ \\
    \textbf{Fundus} & \cellcolor{ired!20}$58.65\pm1.68$ & \cellcolor{ired!40}$66.08\pm1.16$ & \cellcolor{ired!80}$82.79\pm1.83$ & $34.33\pm2.10$ & \cellcolor{ired!40}$68.68\pm2.49$ & \cellcolor{ired!80}$93.81\pm1.41$ & \cellcolor{ired!60}$79.72\pm2.71$ & \cellcolor{ired!40}$65.86\pm2.79$ & $68.74\pm0.69$ \\
    \textbf{OCT} & \cellcolor{ired!80}$80.38\pm1.36$ & \cellcolor{ired!60}$75.02\pm1.05$ & \cellcolor{ired!80}$86.81\pm1.67$ & $36.11\pm2.05$ & \cellcolor{ired!60}$72.21\pm2.41$ & \cellcolor{ired!80}$81.60\pm2.28$ & \cellcolor{ired!100}$99.88\pm0.18$ & \cellcolor{ired!40}$65.32\pm2.79$ & $74.67\pm0.64$ \\
    \textbf{Microscopy} & \cellcolor{ired!40}$60.17\pm1.73$ & \cellcolor{ired!40}$66.59\pm1.16$ & \cellcolor{ired!80}$83.10\pm1.86$ & $36.60\pm2.10$ & \cellcolor{ired!60}$71.29\pm2.45$ & \cellcolor{ired!60}$77.05\pm2.46$ & \cellcolor{ired!80}$80.66\pm2.71$ & \cellcolor{ired!100}$97.66\pm0.86$ & $71.64\pm0.67$ \\
    \midrule
    \textbf{Overall} & $69.89\pm1.59$ & $72.91\pm1.08$ & $84.52\pm1.73$ & $43.91\pm2.14$ & $73.62\pm2.37$ & $80.10\pm2.41$ & $84.18\pm2.47$ & $71.44\pm2.65$ & $72.57\pm0.66$ \\
    \bottomrule
\end{tabular}
    % \begin{tablenotes}
    %     \item CT - Computed Tomography; MRI -  Magnetic Resonance Imaging; US - Ultrasound; Der - Dermoscopy; FP - Fundus Photography.
    %     \item OCT - Optical Coherence Tomography; Micro - Microscopy Images; X-Ray - X-Ray Imaging
    % \end{tablenotes}
    \label{tab:modality_generalize_nothink}
\end{table*}

\begin{table*}[t]
\vspace{-10pt}
    \centering
    \caption{\textbf{Cross-Modality Generalization Performance of Med-R1 with Think-after RL Post-Training.} Accuracy (\%) across eight medical imaging modalities, where rows indicate training modalities and columns test modalities. Darker cell shades indicate higher accuracy for corresponding training-test pairs in each column. The base model is Qwen2.5-VL-3B.
    % The "Overall" column/row reports average performance per training and test domain.
    }

    \setlength{\tabcolsep}{2.5pt}
    \begin{tabular}{p{0.09\linewidth} | *{8}{>{\centering\arraybackslash}p{0.09\linewidth}} | >{\centering\arraybackslash}p{0.09\linewidth} }
    \toprule
    \multicolumn{10}{c}{\textbf{Think-after RL fine-tuned Qwen2.5-VL-3B}} \\
    \midrule
    \diagbox{Train}{Test} & CT & MRI & X-Ray & Ultrasound & Dermoscopy & Fundus & OCT & Microscopy & Overall \\
    \midrule
    \textbf{CT} & \cellcolor{ired!100}$96.64\pm0.62$ & \cellcolor{ired!60}$72.53\pm1.08$ & \cellcolor{ired!80}$85.45\pm1.67$ & $36.31\pm2.03$ & \cellcolor{ired!40}$66.16\pm2.53$ & \cellcolor{ired!60}$79.78\pm2.37$ & \cellcolor{ired!80}$86.20\pm2.30$ & \cellcolor{ired!40}$68.20\pm2.75$ & $73.91\pm0.65$ \\
    \textbf{MRI} & \cellcolor{ired!40}$63.65\pm1.67$ & \cellcolor{ired!100}$98.87\pm0.26$ & \cellcolor{ired!80}$81.11\pm1.92$ & $33.70\pm2.03$ & \cellcolor{ired!60}$71.82\pm2.45$ & \cellcolor{ired!80}$81.51\pm2.32$ & \cellcolor{ired!80}$86.91\pm2.30$ & \cellcolor{ired!40}$67.57\pm2.70$ & $73.14\pm0.67$ \\
    \textbf{X-Ray} & \cellcolor{ired!60}$79.20\pm1.42$ & \cellcolor{ired!60}$72.70\pm1.10$ & \cellcolor{ired!80}$94.61\pm1.08$ & $31.58\pm2.00$ & \cellcolor{ired!60}$70.21\pm2.45$ & \cellcolor{ired!80}$80.60\pm2.32$ & \cellcolor{ired!80}$85.97\pm2.30$ & \cellcolor{ired!40}$68.65\pm2.70$ & $72.94\pm0.66$ \\
    \textbf{Ultrasound} & \cellcolor{ired!20}$59.06\pm1.70$ & \cellcolor{ired!40}$66.23\pm1.15$ & \cellcolor{ired!80}$80.25\pm1.95$ & \cellcolor{ired!100}$99.37\pm0.34$ & \cellcolor{ired!40}$67.38\pm2.53$ & \cellcolor{ired!60}$74.95\pm2.55$ & \cellcolor{ired!60}$78.77\pm2.77$ & \cellcolor{ired!40}$67.93\pm2.75$ & $74.24\pm0.64$ \\
    \textbf{Dermoscopy} & \cellcolor{ired!20}$56.77\pm1.71$ & \cellcolor{ired!40}$66.86\pm1.15$ & \cellcolor{ired!80}$80.37\pm1.95$ & $34.72\pm2.07$ & \cellcolor{ired!80}$89.51\pm1.68$ & \cellcolor{ired!60}$75.05\pm2.55$ & \cellcolor{ired!60}$78.66\pm2.77$ & \cellcolor{ired!40}$67.30\pm2.75$ & $68.65\pm0.67$ \\
    \textbf{Fundus} & \cellcolor{ired!20}$58.01\pm1.68$ & \cellcolor{ired!40}$65.89\pm1.16$ & \cellcolor{ired!80}$81.30\pm1.86$ & $32.84\pm2.03$ & \cellcolor{ired!40}$68.68\pm2.53$ & \cellcolor{ired!80}$92.53\pm1.55$ & \cellcolor{ired!60}$79.36\pm2.77$ & \cellcolor{ired!40}$66.22\pm2.79$ & $68.10\pm0.69$ \\
    \textbf{OCT} & \cellcolor{ired!60}$79.98\pm1.37$ & \cellcolor{ired!60}$75.45\pm1.05$ & \cellcolor{ired!80}$85.51\pm1.73$ & $34.14\pm2.03$ & \cellcolor{ired!60}$71.21\pm2.45$ & \cellcolor{ired!80}$81.69\pm2.32$ & \cellcolor{ired!100}$99.29\pm0.53$ & \cellcolor{ired!40}$66.94\pm2.75$ & $74.28\pm0.65$ \\
    \textbf{Microscopy} & \cellcolor{ired!20}$59.27\pm1.68$ & \cellcolor{ired!40}$64.22\pm1.19$ & \cellcolor{ired!80}$81.11\pm1.89$ & $34.47\pm2.05$ & \cellcolor{ired!40}$63.86\pm2.60$ & \cellcolor{ired!60}$73.68\pm2.64$ & \cellcolor{ired!60}$78.42\pm2.77$ & \cellcolor{ired!80}$86.67\pm2.03$ & $67.71\pm0.69$ \\
    \midrule
    \textbf{Overall} & $69.07\pm1.62$ & $72.84\pm1.10$ & $83.72\pm1.76$ & $42.14\pm2.12$ & $71.10\pm2.53$ & $79.97\pm2.37$ & $84.20\pm2.47$ & $69.93\pm2.65$ & $71.62\pm0.66$ \\
    \bottomrule
\end{tabular}
    \begin{tablenotes}
        \item CT - Computed Tomography; MRI -  Magnetic Resonance Imaging; US - Ultrasound; Der - Dermoscopy; FP - Fundus Photography.
        \item OCT - Optical Coherence Tomography; Micro - Microscopy Images; X-Ray - X-Ray Imaging
    \end{tablenotes}
    \label{tab:modality_Think_after}

\vspace{-10pt}
\end{table*}

\smallskip\noindent\textbf{Reward design:} We follow \cite{shao2024deepseekmath} and use two types of reward: format and accuracy. Firstly, we prompt the model to explicitly output its thinking process in the ``<think>...</think>" tag and the final answer in the ``<answer>...</answer>" tag. The format reward is designed to check if the aforementioned tags are present in the final response. A reward score of 1 will be given if they exist and are correct. This helps the model to organize its thoughts and answer in a structured format for the ease of reading. The accuracy reward is a rule-based reward that checks if the actual answer matches with the ground truth. Similarly, a reward score of 1 is given when the results match. In practice, the ground truths are letter options ``A, B, C, D" for multiple choice questions, and we treat all responses with correct letter options as the leading word as correct (``A...").

\subsection{No-Thinking Med-R1}
% Previous work~\cite{li2025cls} found that for the image classification task, compared with normal rule-based RL fine-tuning, excluding format accuracy and letting the model only output the final answer could lead to better performance.
% However, they only discussed this phenomenon for image classification task. In this paper, we further explore the thinking process of medical VQA and propose the No-Thinking-Med-R1. In contrast to \cite{li2025cls}, which removes format rewards entirely, we remove only the <think> tag, and retain the <answer> tag. 
Previous work~\cite{li2025cls} found that, in image classification tasks, removing both reasoning and format supervision during RL could sometimes improve performance. However, that finding is specific to structured classification tasks with simple output space, where reasoning is rarely required. In contrast, medical VQA involves multimodal, semantically complex inputs, and reasoning failures can arise from domain mismatch rather than verbosity alone~\cite{shao2024deepseekmath,moor2023med}.
Given the difficulty of obtaining reliable CoT annotations in medical settings, we further investigate the role of intermediate reasoning in RL post-training.
We revise the instruction prompt as \texttt{\{Question\}. Output the single-letter choice (A, B, C, D,...) in <answer>...</answer> tags.}, where \texttt{\{Question\}} will be replaced by each specific question. By doing so, we encourage the model only to output the final answer without any explicit thinking process. The accuracy reward is maintained, where the reward score is 1 when the extracted answer matches the ground truth labels. Note that when there is any text outside the <answer> tag, i.e., an explicit thinking process exists, the extracted content will be null, and therefore the accuracy reward will be 0. Therefore, the model will be forced to generate only the answers.

\begin{table*}[t]
    \centering
    \caption{\textbf{Performance comparison across eight medical modalities.} Our GRPO-finetuned model outperforms zero-shot general-purpose VLMs, medical-domain VLMs, and supervised fine-tuning baselines, while maintaining scalability. The best and second-best performances per column are highlighted in {\color{i1}red} and {\color{i2}blue}. Modality abbreviations: CT – Computed Tomography; MRI – Magnetic Resonance Imaging; US – Ultrasound; Der – Dermoscopy; FP – Fundus Photography; OCT – Optical Coherence Tomography; Micro – Microscopy; X-Ray – X-ray Imaging.
}
    \begin{tabular}{l|cccccccc|c}
        \toprule
       \diagbox{Methods}{Modality} & CT & MRI & X-Ray & Ultrasound & Dermoscopy & Fundus & OCT & Microscopy & Overall \\
        \midrule
         \multicolumn{10}{c}{\textbf{Zero-shot VLMs}} \\
        \midrule
        $\text{BLIP-2}^{\dagger}$~\cite{li2023blip} &56.74 &41.32&67.58&37.27&40.65&46.24&68.08&50.40& 51.04\\
        $\text{InstructBLIP}^{\dagger}$~\cite{dai2023instructblip} & 28.72 &33.15&61.04&41.25&62.22&50.31&42.59&46.29&45.70\\
        $\text{LLaVA}^{\dagger}$~\cite{liu2023visual} & 17.73 & 26.72 & 30.70 & 18.66 & 49.74 & 47.11 & 33.73 & 28.87 & 31.66\\
        $\text{LLaMA Adapter v2}^{\dagger}$~\cite{gao2023llama} & 21.41 & 26.63 & 46.44 & 34.05 & 51.76 & 50.74 & 33.00 & 38.66 & 37.83\\
        $\text{MiniGPT-4}^{\dagger}$~\cite{zhu2023minigpt} & 22.81 & 27.48 & 38.30 & 25.50 & 40.25 & 38.33 & 31.40 & 36.23 & 32.54\\
        InternVL2~\cite{chen2024expanding} & 40.20 & 58.10 & 57.90 & 49.10 & 51.90 & 53.20 & 59.10 & 64.00 & 54.19\\
        Qwen2-VL-2B~\cite{bai2023qwen} & 45.10 & 38.57 & 39.32 & 30.86 & 35.83 & 43.17 & 35.14 & 36.85  & 38.11 \\
        Qwen2-VL-7B~\cite{bai2023qwen} & 61.46 & 45.77 & 64.27 & 36.01 & 49.08 & 59.84 & 59.32 & 61.08  & 54.60 \\
        Qwen2-VL-72B~\cite{wang2024qwen2} & 67.97 &  69.39 &  77.21 & 51.39 & 65.31 & 72.58 & 72.76 & 67.83 & 68.05\\
        Qwen2.5-VL-3B~\cite{bai2025qwen2} & 53.87 & 54.23 & 61.84 & 32.69 & 52.94 & 62.47 & 56.23 & 59.64 & 54.24\\
        Qwen2.5-VL-7B~\cite{bai2025qwen2} & 60.44 & 58.44 & 73.99 & 30.66 & 62.48 & 67.30 & 61.20 & 67.84 & 60.29\\
        Qwen2.5-VL-72B~\cite{bai2025qwen2} & 66.18 & 68.74 & 77.59 & 49.81 & 69.75 & 71.04 & 69.22 & 69.37 & 67.71\\
        
        \midrule

        \multicolumn{10}{c}{\textbf{Zero-shot Medical VLMs}} \\
        \midrule
        $\text{LLaVA-Med}^{\dagger}$~\cite{li2023llava} & 18.69 &27.47 &30.68&29.88&44.95&39.03&34.61&33.29&32.33\\
        $\text{RadFM}^{\dagger}$~\cite{wu2023towards} & 27.56 & 24.06 & 30.95 & 16.57 & 39.21 & 36.89 & 32.80 & 27.97 & 29.50\\
        $\text{Med-Flamingo}^{\dagger}$~\cite{moor2023med} &38.47&40.56&30.34&24.64&32.43&30.12&26.51&19.93&30.38\\
        $\text{MedVInT}^{\dagger}$~\cite{zhang2023pmc}&40.74&43.10&55.10&41.26&29.11&31.84&23.26&32.00&37.05\\
        HuatuoGPT-Vision~\cite{chen2024huatuogpt} & 35.30 & 40.40 & 41.50 & 60.10 & 53.10 & 51.40 & 59.30 & 62.30 & 50.43\\
        HealthGPT~\cite{lin2025healthgpt} & 35.50 & 78.50 & 81.90 & 51.40 & 64.90 & 54.60 & 89.30 & 88.20 & 68.04\\
        
        % Huatuo-GPT-vision~\cite{chen2024huatuogpt}\\
        
        \midrule
        \multicolumn{10}{c}{\textbf{Fine-tuned VLMs}} \\
        \midrule
        Qwen2-VL-2B (SFT) & 51.74 & 52.83 & 65.57 & 47.65 & 51.91 & 52.26 & 53.99 & 56.58 & 54.07\\
        Qwen2.5-VL-3B (SFT) & 56.06 & 60.81 & 69.23 & 41.77 & 60.11 & 69.19 & 63.95 & 65.66 & 60.85\\
        \textbf{Qwen2-VL-2B (Think)} &  66.30 &  71.67 &  77.73 & \cellcolor{i1!30}57.31 & 72.33 & 71.20 & 71.96 & 70.80  & 69.91 \\
        \textbf{Qwen2-VL-2B (Nothink) } & \cellcolor{i2!30}72.19 & \cellcolor{i1!30}74.37 & 78.37 & \cellcolor{i2!30}54.43 & \cellcolor{i1!30}74.73 & 75.07 & 76.59 & \cellcolor{i1!30}74.13 & 72.49\\
        \textbf{Qwen2.5-VL-3B (Think)} & \cellcolor{i1!30} 73.46 & 68.77 & \cellcolor{i2!30}83.72 & 45.21 & 68.88 & 77.40 & 82.18 & 69.47 & 71.14\\
        \textbf{Qwen2.5-VL-3B (Think-after)} & 69.07 & 72.84 & \cellcolor{i2!30}83.72 & 42.14 & 71.10 & \cellcolor{i2!30}79.97 & \cellcolor{i1!30}84.20 & 69.93 & 71.62\\
        \textbf{Qwen2.5-VL-3B (Nothink)} & 
        69.89 & \cellcolor{i2!30}72.91 & \cellcolor{i1!30} 84.52 & 43.91 & \cellcolor{i2!30}73.62 &  \cellcolor{i1!30}80.10 & \cellcolor{i2!30}84.18 & \cellcolor{i2!30}71.44 & \textbf{72.57}\\
        \bottomrule
    \end{tabular}

    \label{tab:modality_comparation}
\end{table*}

\subsection{Think-after Med-R1}
As said in the previous section, removing both reasoning and format supervision during RL could sometimes improve performance. However, as medical AI systems must not only achieve high accuracy but also provide reasoning that physicians can review and validate.
To address this, we introduce a new reasoning protocol termed \textbf{Think-After}, in which the model first predicts the answer and then generates a post-hoc rationale explaining that decision. This design preserves interpretability while minimizing the instability introduced by lengthy reasoning chains. Concretely, the instruction prompt is revised as:
\texttt{\{Question\}. Output the single-letter choice (A, B, C, D,...) in <answer>...</answer> tags. Then provide the reasoning step by step to explain why you chose this answer.}
where \texttt{\{Question\}} is replaced by each specific VQA item. By separating answer prediction and rationale generation, we reduce interference between reasoning and decision processes, achieving balanced trade-off between accuracy and interpretability.

\section{Experiment \& Results}
% This section introduces our experimental setup, implementation details, and results.

\subsection{Setup}
\smallskip\noindent\textbf{Datasets.}
We adopt the VQA data from the open-access part of the OmniMedVQA benchmark~\cite{hu2024omnimedvqa}, which consists of a total of $82,059$ images and $88,996$ vision question answering pairs. 
OmniMedVQA includes VQA pairs from eight imaging modalities: CT ($15,808$), MRI ($31,877$), X-Ray ($7,916$), Ultrasound ($10,991$), Dermoscopy ($6,679$), Fundus ($5,398$), OCT ($4,646$), and Microscopy ($5,680$). 
It is also categorized into five VQA question types, including Anatomy Identification ($16,448$), Disease Diagnosis ($55,387$), Lesion Grading ($2,098$), Modality Recognition ($11,565$), and Other Biological Attributes ($3,498$). We split the dataset into training and test sets following an 80-20 ratio for each setting.
% CT ($15,808$)~\cite{chest_ct_scan,covid_ct,sars_cov_2_ct_scan,radimagenet}, 
% MRI ($31,877$)~\cite{breakhis,nlm_malaria_data,radimagenet}, 
% X-Ray ($7,916$)~\cite{oct_x_ray_2017,knee_osteoarthritis,mura,covidx_cxr_4}, 
% Ultrasound ($10,991$)~\cite{radimagenet}, 
% Dermoscopy ($6,679$)~\cite{fitzpatrick_17k,isbi2016,isic2019,monkeypox_skin_image_2022,pad_ufes_20}, 
% Fundus ($5,398$)~\cite{acrima,adam_challenge,diabetic_retinopathy,jsiec,olives}, 
% OCT ($4,646$)~\cite{oct_x_ray_2017,retinal_oct_c8}, 
% and Microscopy ($5,680$)~\cite{breakhis,nlm_malaria_data}.

\smallskip\noindent\textbf{Implementation Details}
\label{sec:implementation}
Training is conducted on HGX H100~\cite{choquette2023nvidia} server with 2×H100 GPUs (80GB VRAM) using PyTorch~\cite{paszke2019pytorch} and FlashAttention-2~\cite{dao2023flashattention} for optimized efficiency. We initialize from Qwen2-VL-2B-Instruct~\cite{bai2023qwen} with full parameter tuning, employing per-GPU batch size 1 (effective batch size 4 via 2-step gradient accumulation) and bfloat16 mixed precision. Input sequences combine visual embeddings from 328×328 resolution images (max 401k pixels) with textual prompts truncated to 1,024 tokens. The GRPO policy generates four candidate rationales per sample, with a sampling temperature of $\tau=0.7$. Each training is run for one epoch.

\smallskip\noindent\textbf{Task setting.} We evaluate our approach in two distinct generalization settings using the OmniMedVQA dataset \cite{hu2024omnimedvqa}: cross-modality generalization and cross-task generalization.  
\begin{itemize}
    \item \textbf{Cross-modality generalization}: We train our model on a single modality at a time (out of 8 available modalities) and evaluate its performance on the other seven modalities.
    \item \textbf{Cross-task generalization}: We identify 5 distinct task types within the dataset and adopt the same train-test partitioning strategy as in the cross-modality setting, training on one task and evaluating on the other four.
\end{itemize}

\begin{table*}[t]
\vspace{-15pt}
    \centering
    \caption{\textbf{Cross-Task Generalization of Med-R1:} Performance is evaluated across five clinical reasoning task types (rows: training tasks, columns: test tasks), with darker cell shading indicating stronger generalization. Results demonstrate that domain-specific training (e.g., Disease Diagnosis) preserves in-task expertise while maintaining adaptability to unseen tasks, particularly for modality-agnostic skills like Modality Recognition. The base model is Qwen2.5-VL-3B.}
    \begin{tabular}{c|ccccc|c}
    \toprule
    \multicolumn{7}{c}{\textbf{RL fine-tuned Qwen2.5-VL-3B}} \\
    \midrule
    \multirow{2}{*}{\diagbox{Train}{Test}} & Anatomy & Disease & Lesion & Modality & Other & \multirow{2}{*}{Overall} \\
    & Identification & Diagnosis & Grading & Recognition & Attributes \\
    \midrule
    Anatomy Identification & \cellcolor{ired!80}$94.65\pm0.75$ & \cellcolor{ired!40}$61.87\pm0.90$ & \cellcolor{ired!40}$63.30\pm4.59$ & \cellcolor{ired!100}$96.66\pm0.73$ & \cellcolor{ired!60}$78.69\pm3.05$ & $79.04\pm0.65$ \\
    Disease Diagnosis & $43.83\pm1.67$ & \cellcolor{ired!100}$97.74\pm0.28$ & \cellcolor{ired!60}$79.82\pm3.67$ & \cellcolor{ired!100}$96.66\pm0.73$ & \cellcolor{ired!80}$87.78\pm2.41$ & $81.16\pm0.49$ \\
    Lesion Grading & $47.10\pm1.71$ & \cellcolor{ired!20}$59.33\pm0.93$ & \cellcolor{ired!80}$88.99\pm2.98$ & \cellcolor{ired!100}$96.03\pm0.77$ & \cellcolor{ired!40}$69.03\pm3.41$ & $72.10\pm0.70$ \\
    Modality Recognition & $45.96\pm1.67$ & \cellcolor{ired!20}$58.63\pm0.91$ & \cellcolor{ired!40}$67.89\pm4.36$ & \cellcolor{ired!100}$99.08\pm0.40$ & \cellcolor{ired!40}$69.46\pm3.41$ & $68.20\pm0.71$ \\
    Other Attributes & $44.04\pm1.68$ & \cellcolor{ired!40}$60.09\pm0.92$ & \cellcolor{ired!40}$60.09\pm4.59$ & \cellcolor{ired!100}$95.74\pm0.82$ & \cellcolor{ired!80}$94.74\pm1.63$ & $70.94\pm0.70$ \\
    \midrule
    \textbf{Overall} & $55.12\pm0.75$ & $67.53\pm0.39$ & $72.02\pm1.86$ & $96.83\pm0.31$ & $79.94\pm1.32$ & $74.29\pm0.65$ \\
    \bottomrule
\end{tabular}
\vspace{-10pt}
    \label{tab:medical_reasoning}
\end{table*}

\begin{table*}[t]
% \vspace{-10pt}
    \centering
    \caption{\textbf{No-Think Cross-Task Generalization of Med-R1:} Performance is evaluated across five clinical reasoning task types (rows: training tasks, columns: test tasks), with darker cell shading indicating stronger generalization.}
    \begin{tabular}{c|ccccc|c}
    \toprule
    \multicolumn{7}{c}{\textbf{No-Think RL fine-tuned Qwen2.5-VL-3B}} \\
    \midrule
    \multirow{2}{*}{\diagbox{Train}{Test}} & Anatomy & Disease & Lesion & Modality & Other & \multirow{2}{*}{Overall} \\
    & Identification & Diagnosis & Grading & Recognition & Attributes \\
    \midrule
    Anatomy Identification & \cellcolor{ired!100}$96.85\pm0.57$ & \cellcolor{ired!40}$63.86\pm0.90$ & \cellcolor{ired!60}$79.59\pm3.78$ & \cellcolor{ired!100}$96.82\pm0.69$ & \cellcolor{ired!60}$77.41\pm3.12$ & $82.91\pm0.64$ \\
    Disease Diagnosis & $44.82\pm1.73$ & \cellcolor{ired!100}$98.78\pm0.21$ & \cellcolor{ired!80}$85.78\pm3.33$ & \cellcolor{ired!100}$95.82\pm0.82$ & \cellcolor{ired!80}$90.91\pm2.13$ & $83.22\pm0.48$ \\
    Lesion Grading & $44.61\pm1.70$ & \cellcolor{ired!40}$60.27\pm0.91$ & \cellcolor{ired!100}$97.02\pm1.61$ & \cellcolor{ired!100}$95.90\pm0.79$ & \cellcolor{ired!80}$82.10\pm2.84$ & $75.98\pm0.70$ \\
    Modality Recognition & $47.07\pm1.71$ & \cellcolor{ired!20}$59.58\pm0.91$ & \cellcolor{ired!60}$75.92\pm3.90$ & \cellcolor{ired!100}$99.75\pm0.19$ & \cellcolor{ired!80}$81.53\pm2.84$ & $72.77\pm0.70$ \\
    Other Attributes & $43.98\pm1.71$ & \cellcolor{ired!40}$61.11\pm0.90$ & \cellcolor{ired!40}$61.70\pm4.59$ & \cellcolor{ired!100}$96.07\pm0.77$ & \cellcolor{ired!100}$95.60\pm1.49$ & $71.69\pm0.71$ \\
    \midrule
    \textbf{Overall} & $55.46\pm0.77$ & $68.72\pm0.39$ & $80.00\pm1.67$ & $96.87\pm0.31$ & $85.51\pm1.16$ & $77.31\pm0.62$ \\
    \bottomrule
\end{tabular}
\vspace{-10pt}
    \label{tab:no_thinking_medical_reasoning}
\end{table*}

We focus on VQA, which integrates core vision–language abilities such as classification, grounding, and reasoning into a unified framework, making it a representative precursor to detection or captioning tasks.

\smallskip\noindent\textbf{Baseline Methods \& Evaluation Metric.}

We report our results by separating the baselines into three groups. Zero-shot VLMs are models pre-trained for general-purpose VQA without medical adaptation. Medical VLMs are models pre-trained on medical data specifically for medical VQA. Fine-tuned VLMs are models trained on OmniMedVQA using supervised fine-tuning and RL fine-tuning.
We evaluate model performance using \textbf{VQA choice accuracy}, the standard metric for medical VQA, where the model selects the correct answer from $K$ clinically validated options.  
Given an image $I$, a question $Q$, and candidate answers $\{A_k\}_{k=1}^{K}$, accuracy is defined as:
\begin{equation}
\text{Accuracy} = \frac{1}{N}\sum_{i=1}^{N} \mathbb{I}(\hat{y}_i = y_i),
\end{equation}
where $N$ is the total number of test cases, $\hat{y}_i$ is the predicted answer index, $y_i$ the ground truth, and $\mathbb{I}$ the indicator function (1 if correct, 0 otherwise).

To assess result reliability, we computed \textbf{95\% bootstrap confidence intervals (CIs)} for all model variants across modalities and tasks. Each experiment performed 10,000 bootstrap resamplings of binary accuracy labels to estimate the mean and confidence bounds, reporting $mean \pm half\text{-}width$ for each model.

All fine-tuned VLM results (SFT, GRPO+Think, GRPO+Think-After, and GRPO+No-Think) are computed under the identical averaging protocol—macro-averaged accuracy across all cross-modality training–testing pairs—to ensure fair and consistent comparison.

\subsection{Cross-Modality Generalization}
\label{Cross-Modality Generalization}
	We comprehensively evaluate Med-R1’s adaptability across \textbf{eight} medical imaging modalities, including Computed Tomography, Magnetic Resonance Imaging, Ultrasound, Dermoscopy, Fundus Photography, Optical Coherence Tomography, Microscopy Images, and X-ray Imaging. Our experiments focus on two key aspects: (1) cross-modal generalization, where the model is trained on one modality and tested on another, and
(2) comparative performance against other popular VLMs and medical-specific VLMs evaluated using zero-shot and SFT.

\begin{table*}[t]
    \centering
    \caption{\textbf{Think-After Cross-Task Generalization of Med-R1:} Performance is evaluated across five clinical reasoning task types (rows: training tasks, columns: test tasks), with darker cell shading indicating stronger generalization.}
    \begin{tabular}{c|ccccc|c}
    \toprule
    \multicolumn{7}{c}{\textbf{Think-After fine-tuned Qwen2.5-VL-3B}} \\
    \midrule
    \multirow{2}{*}{\diagbox{Train}{Test}} & Anatomy & Disease & Lesion & Modality & Other & \multirow{2}{*}{Overall} \\
    & Identification & Diagnosis & Grading & Recognition & Attributes \\
    \midrule
    Anatomy Identification & \cellcolor{ired!100}$95.22\pm0.74$ & \cellcolor{ired!40}$60.62\pm0.90$ & \cellcolor{ired!40}$65.83\pm4.36$ & \cellcolor{ired!100}$96.57\pm0.73$ & \cellcolor{ired!80}$81.53\pm2.91$ & $79.95\pm0.65$ \\
    Disease Diagnosis & $43.11\pm1.68$ & \cellcolor{ired!100}$97.86\pm0.27$ & \cellcolor{ired!80}$81.65\pm3.67$ & \cellcolor{ired!100}$95.99\pm0.77$ & \cellcolor{ired!80}$92.33\pm1.99$ & $82.19\pm0.49$ \\
    Lesion Grading & $43.56\pm1.70$ & \cellcolor{ired!20}$59.78\pm0.92$ & \cellcolor{ired!80}$85.55\pm3.21$ & \cellcolor{ired!100}$96.20\pm0.77$ & \cellcolor{ired!60}$78.55\pm3.05$ & $72.73\pm0.71$ \\
    Modality Recognition & $44.73\pm1.71$ & \cellcolor{ired!20}$58.82\pm0.93$ & \cellcolor{ired!40}$68.12\pm4.47$ & \cellcolor{ired!100}$98.87\pm0.44$ & \cellcolor{ired!60}$78.12\pm3.05$ & $69.73\pm0.71$ \\
    Other Attributes & $45.21\pm1.67$ & \cellcolor{ired!20}$59.65\pm0.90$ & \cellcolor{ired!40}$67.43\pm4.36$ & \cellcolor{ired!100}$95.74\pm0.82$ & \cellcolor{ired!100}$95.74\pm1.49$ & $72.75\pm0.71$ \\
    \midrule
    \textbf{Overall} & $54.36\pm0.75$ & $67.34\pm0.40$ & $73.72\pm1.86$ & $96.67\pm0.32$ & $85.26\pm1.18$ & $75.47\pm0.63$ \\
    \bottomrule
\end{tabular}
    \label{tab:q_type_think_after}
\vspace{-10pt}
\end{table*}

\smallskip\noindent\textbf{Results on generalization.} To evaluate Med-R1’s cross-modality generalization, we measure its accuracy across eight distinct medical imaging modalities (Table~\ref{tab:modality_generalize}). Overall row and column summarize the model’s average performance across training and test domains, providing insights into its generalization ability.
Med-R1 achieves a strong overall accuracy of 69.91\%, demonstrating its ability to generalize across diverse medical imaging modalities. Notably, models trained on CT, MRI, and X-Ray exhibit the highest generalization capability, with overall scores of 71.44\%, 71.26\%, and 72.35\%, respectively. In contrast, models trained on Fundus Photography and Microscopy images show lower generalization, with 67.67\% and 67.54\% overall accuracy, indicating that certain modality-specific features (e.g., texture-based imaging in US and Micro) may not transfer as effectively to other domains.
Importantly, the overall test accuracy of 69.91\% highlights Med-R1’s ability to perform well across unseen imaging modalities, despite being trained on a single domain at a time. This result underscores the effectiveness of reinforcement learning in enhancing cross-modality transfer, allowing the model to maintain robust performance without requiring extensive retraining for each medical imaging modality.
% These findings suggest that RL-based fine-tuning improves domain adaptation, enabling medical VLMs to generalize beyond their training modality, a crucial step toward developing clinically deployable and scalable vision-language models for diverse medical applications.

\begin{table*}[t]
    \centering
    \caption{\textbf{Performance Comparison of VLMs on Five Medical VQA Tasks:} GRPO Fine-Tuning Outperforms Zero-Shot and SFT Baselines Across Diverse Reasoning Tasks. Performance is evaluated on five clinical reasoning types (columns) across three model categories: general-purpose VLMs (zero-shot), medical VLMs (zero-shot), and fine-tuned VLMs. The best and second-best performances are marked in red and blue.}
    \begin{tabular}{l|ccccc|c}
        \toprule
       \multirow{2}{*}{\diagbox{Methods}{Types}} & Anatomy& Disease& Lesion & Modality& Other& \multirow{2}{*}{Overall} \\
        & Identification & Diagnosis & Grading & Recognition &Attributes \\
        \midrule
         \multicolumn{7}{c}{\textbf{Zero-shot VLMs}} \\
        \midrule
        $\text{BLIP-2}^{\dagger}$~\cite{li2023blip}& 44.39 & 44.51 & 29.03 & 68.19 & 67.95 & 48.12\\
        $\text{InstructBLIP}^{\dagger}$~\cite{dai2023instructblip}& 44.35 & 32.29 & 59.25 & 75.27 & 23.72 & 40.40\\
        $\text{LLaVA}^{\dagger}$~\cite{liu2023visual} & 25.86 & 29.10 & 43.95 & 21.36 & 31.90 & 27.96\\
        $\text{LLaMA Adapter v2}^{\dagger}$~\cite{gao2023llama}& 33.72 & 31.19 & 41.99 & 37.29 & 34.22 & 32.82\\
        $\text{MiniGPT-4}^{\dagger}$~\cite{zhu2023minigpt} & 28.88 & 30.47 & 34.56 &  26.43 & 30.36 & 29.74 \\
        Qwen2-VL-2B~\cite{bai2023qwen} & 30.70 & 36.53 & 43.58 & 59.90 & 42.19 & 42.58\\
        Qwen2-VL-7B~\cite{bai2023qwen} & 42.57 & 48.83 & 52.06 & 84.74 & 59.66 & 57.57\\
        Qwen2-VL-72B~\cite{wang2024qwen2} & 56.41 & 65.71 & 62.15 & 98.11 & 80.53 & 72.58\\
        Qwen2.5-VL-3B~\cite{bai2025qwen2} & 35.23 & 50.45 & 52.79 & 85.23 & 54.77 & 55.69\\
        Qwen2.5-VL-7B~\cite{bai2025qwen2} & 41.00 & 61.32 & 54.13 & 97.78 & 70.45 & 64.94\\
        Qwen2.5-VL-72B~\cite{bai2025qwen2} & 57.22 & 62.55 & 60.32 & 98.20 & 77.41 & 71.14\\
        \midrule

        \multicolumn{7}{c}{\textbf{Zero-shot Medical VLMs}} \\
        \midrule
        $\text{LLaVA-Med}^{\dagger}$~\cite{li2023llava} & 29.53 & 29.22 & 34.18 & 26.93 & 33.08 & 29.25 \\
        $\text{RadFM}^{\dagger}$~\cite{wu2023towards} & 13.31 & 21.69 & 30.35 & 26.64 & 43.85 & 26.99 \\
        $\text{Med-Flamingo}^{\dagger}$~\cite{moor2023med} & 24.93 & 38.90 & 30.74 & 30.19 & 14.18 & 34.03\\
        $\text{MedVInT}^{\dagger}$~\cite{zhang2023pmc}&40.26 & 35.78 & 12.77 & 68.10 & 30.30 & 40.04\\
        
        % Huatuo-GPT-vision~\cite{chen2024huatuogpt}\\
        
        \midrule
        \multicolumn{7}{c}{\textbf{Fine-tuned VLMs}} \\
        \midrule
        Qwen2-VL-2B (SFT) & 53.97 & 51.62 & 60.71 & 86.77 & 63.91 & 63.39\\
        Qwen2.5-VL-3B (SFT) & 54.91 & 57.75 & 64.04 & 84.95 & 71.56 & 66.64\\
        \textbf{Qwen2-VL-2B (Think)} & \cellcolor{i2!30} 62.88	& 66.08	& 65.87	& \cellcolor{i2!30} 98.24	& 80.14 &  74.64\\
        \textbf{Qwen2-VL-2B (Nothink)} & \cellcolor{i1!30} 63.74 & 66.32 & 66.33 & \cellcolor{i1!30} 98.44 &  81.31 & 75.22\\
        \textbf{Qwen2.5-VL-3B (Think)} & 55.12 & \cellcolor{i2!30} 67.53 &  72.02 & 96.83 & 79.94 & 74.29 \\
        \textbf{Qwen2.5-VL-3B (Think-after)} & 54.36 & 67.34 & \cellcolor{i2!30} 73.72 & 96.67 & \cellcolor{i2!30} 85.26 & 75.47 \\
        \textbf{Qwen2.5-VL-3B (Nothink)} & 55.47 & \cellcolor{i1!30} 68.72 & \cellcolor{i1!30} 80.00 & 96.87 & \cellcolor{i1!30} 85.51 & \textbf{77.31}\\
        \bottomrule
    \end{tabular}
    
    \label{tab:question_type_comparation}
    \vspace{-10pt}
\end{table*}

\smallskip\noindent\textbf{Comparison to zero-shot and SFT evaluations with other VLMs.}
As demonstrated in Table~\ref{tab:modality_comparation}, Med-R1 demonstrates its superiority across all eight medical imaging modalities while maintaining exceptional parameter efficiency. For zero-shot results, each cell denotes the zero-shot evaluation accuracy of the model on the particular modality. For all the fine-tuned VLM results, each cell represents the overall accuracy, reflecting the average generalization performance of the given modality when evaluated using models that were separately fine-tuned on each of the eight training modalities. Against general-purpose VLMs, our 2B-parameter model achieves 69.91\% overall accuracy, surpassing the 72B-parameter Qwen2-VL by 1.86\%—a notable result given the \textbf{$36\times$} parameter disparity. This advantage amplifies in critical diagnostic tasks: Med-R1 attains 71.67\% accuracy in MRI compared to Qwen2-VL-72B's 69.39\%, and achieves 72.33\% versus 65.31\% in dermoscopy, challenging the prevailing scale-equals-performance paradigm.
The limitations of specialized medical VLMs become evident through Med-Flamingo's 30.38\% average accuracy, which Med-R1 outperforms by 39.53\%. This stark contrast underscores the ineffectiveness of narrow medical pretraining compared to our RL-driven adaptation strategy. When compared to supervised fine-tuning approaches, GRPO delivers 15.84\% accuracy gains over SFT-tuned Qwen2-VL-2B (69.91\% vs. 54.07\%), with particularly significant improvements in CT interpretation (66.30\% vs. 51.74\%) and OCT analysis (71.96\% vs. 53.99\%).

\begin{figure*}[h]
\vspace{-10pt}
	\centering
\includegraphics[width=0.95\linewidth]{reward_compare.png}
	\caption{\textbf{Training accuracy–reward curves of the Think, No-Think, and Think-After models on CT and MRI datasets.}}
    
    \label{reward_compare}
    \vspace{-10pt}
\end{figure*}
\subsection{Cross-Task Generalization}
We also evaluate Med-R1's generalization across \textbf{five} important clinical tasks \cite{hu2024omnimedvqa}: Anatomy Identification, Disease Diagnosis, Lesion Grading, Modality Recognition, and Other Attributes. Similar to \autoref{Cross-Modality Generalization}, we focus our evaluation on two aspects: cross-modality generalization and comparison against SFT and zero-shot with other VLMs.

\begin{figure}[t]
	\centering
\includegraphics[width=1\linewidth]{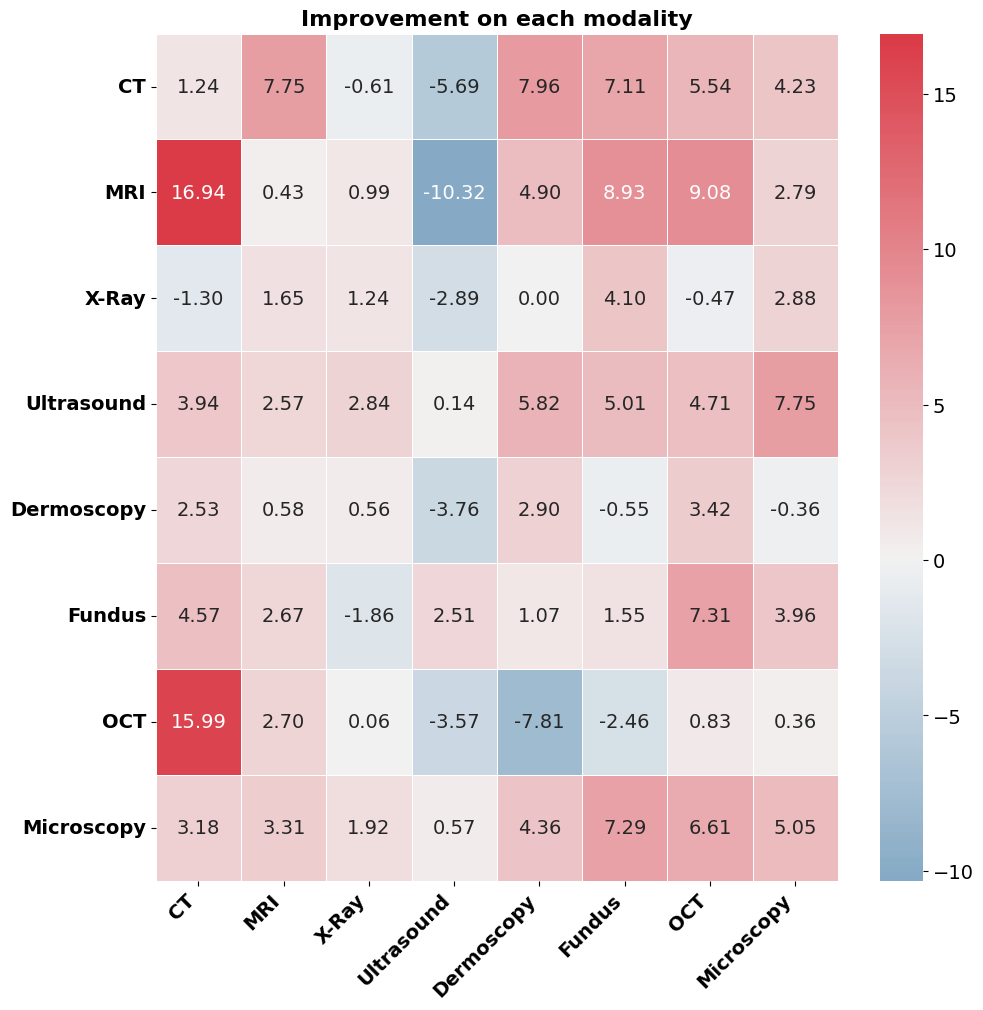}
	\caption{\textbf{Cross-modality accuracy difference between No-Thinking-Med-R1 and Med-R1.} Each cell shows the performance gap (\%) when trained on one modality (y-axis) and evaluated on others (x-axis). Red indicates improvement; blue indicates degradation. 
    % No-Thinking can match or exceed Med-R1 performance, challenging the assumption that reasoning always helps.
    }
    
    \label{fig_nothink_think_modality}
    \vspace{-10pt}
\end{figure}

\smallskip\noindent\textbf{Results on generalization.} As shown in \autoref{tab:medical_reasoning}, models trained on ``disease diagnosis" data achieve the best generalization, with 81.64\% overall accuracy. This suggests that disease diagnosis encompasses diverse feature representations that transfer well across tasks, likely due to its reliance on both anatomical and pathological cues. In contrast, models trained on ``modality recognition" exhibit strong generalization in task-agnostic settings (98.24\% in the ``modality recognition" column), indicating that learning modality distinctions aids in extracting transferable image features. However, training on ``lesion grading" leads to high in-task performance (86.24\%) but relatively lower transferability, implying that this task captures more specialized features that do not generalize as effectively. These results highlight the trade-off between specialization and adaptability, emphasizing the importance of task selection when designing models for broad medical applications.

\begin{figure}[t]
	\centering
\includegraphics[width=1\linewidth]{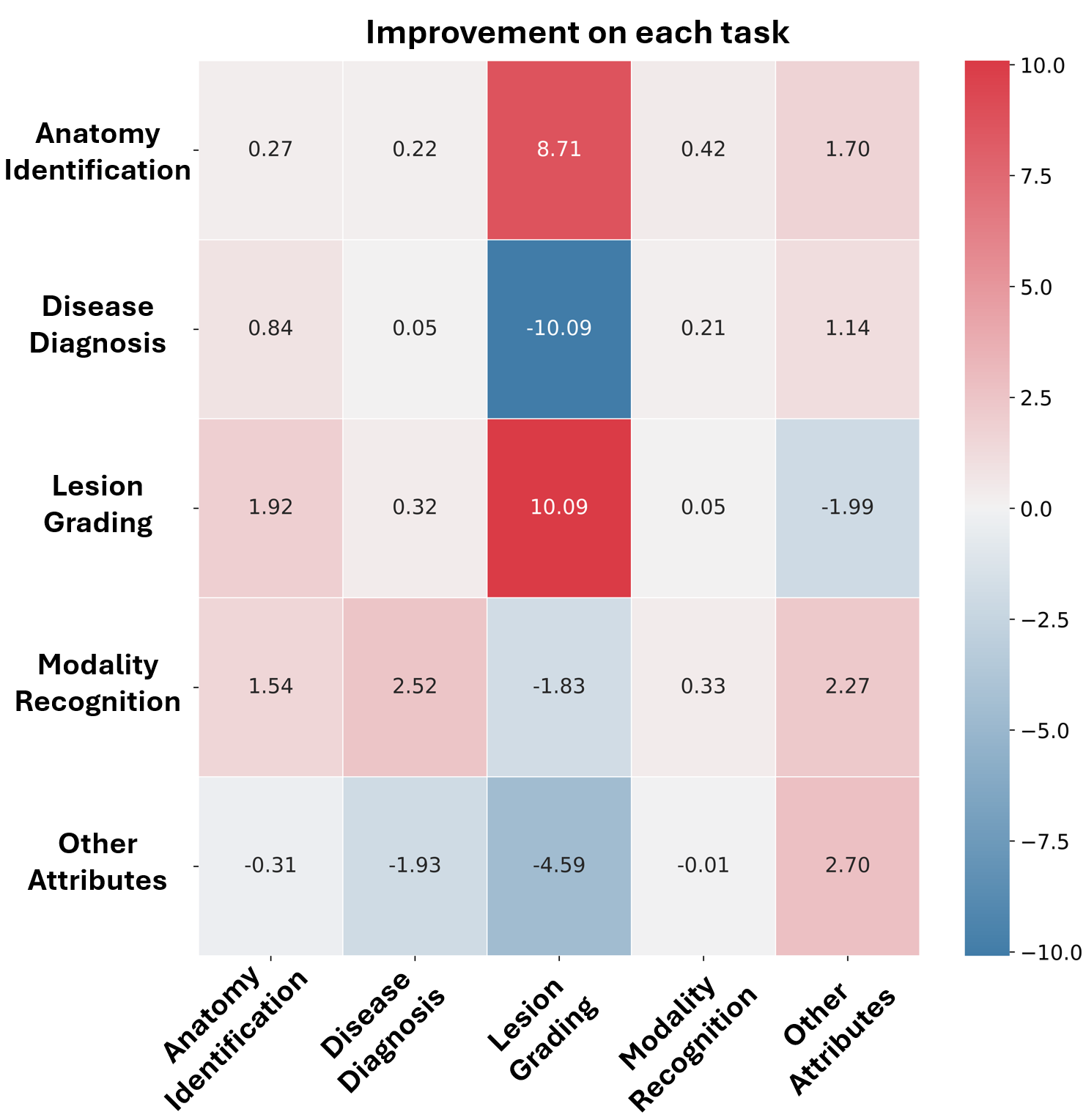}
	\caption{\textbf{Cross-task accuracy difference between No-Thinking-Med-R1 and Med-R1.} Each cell shows the performance gap (\%) when trained on one task (y-axis) and tested on another (x-axis). Red indicates improvement; blue indicates degradation. 
    % No-Thinking often matches or exceeds Med-R1, challenging the assumption that explicit reasoning always improves performance.
    }
    
    \label{fig_nothink_think_task}
    \vspace{-10pt}
\end{figure}

\smallskip\noindent\textbf{Comparison to zero-shot and SFT evaluations with other VLMs.} \autoref{tab:question_type_comparation} shows the comparison results with other popular VLMs evaluated with zero-shot and SFT. For zero-shot results, each cell denotes the zero-shot evaluation accuracy of the model on the particular task. For the fine-tuned VLM results (last two rows), each cell represents the overall accuracy, reflecting the average generalization performance of the given task when evaluated using models that were separately fine-tuned on each of the five training tasks. First of all, the results clearly show that Med-R1 outperforms all other popular VLMs' zero-shot generalization. Remarkably, Med-R1 even outperforms Qwen2-VL-72B (74.64\% vs. 72.58\%), a model with \textbf{70 billion} more parameters. More importantly, this suggests that RL can effectively elevate small models with moderate capacity, opening doors for many real-world applications where resource is a constraint. In contrast, the average generalization with the identical base model trained with SFT is merely 63.39\%, 11.25\% below Med-R1, further demonstrating the strong generalization capability of Med-R1.

\subsection{Analysis of No-Think}
\label{nothink}
% We present the results of No-Thinking-Med-R1 in this section. The results of cross-task are shown in \autoref{tab:no_thinking_medical_reasoning}, and the results of cross-modality are shown in \autoref{tab:modality_generalize_nothink}. Overall, No-Thinking-Med-R1 shows better in-domain results among all settings compared with Med-R1, which demonstrates that No-Thinking-Med-R1 could lead to better in-domain learning results. We then compare the generalization ability of Med-R1 and No-Thinking-Med-R1. For better understanding, we visualize the improvement of No-Thinking-Med-R1 than Med-R1 in Figure~\ref{fig_nothink_think_modality} and Figure~\ref{fig_nothink_think_task}. In cross-modality scenarios, No-Thinking-Med-R1 consistently exhibited superior generalization compared to Med-R1. However, in cross-task settings, No-Thinking-Med-R1's generalization performance varied, showing improvement in some cases and degradation in others. This indicates that No-Thinking-Med-R1 enhances cross-modality generalization, while both methods demonstrate distinct advantages in cross-task generalization. 
We present the results of No-Thinking-Med-R1 in this section. Cross-task performance is shown in \autoref{tab:no_thinking_medical_reasoning}, and cross-modality performance is shown in \autoref{tab:modality_generalize_nothink}. Overall, No-Thinking-Med-R1 achieves stronger in-domain performance across all settings compared to Med-R1, suggesting that removing explicit reasoning generation may lead to more effective task-specific learning.
To assess generalization, we compare Med-R1 and No-Thinking-Med-R1 across modalities and tasks. As visualized in Figure~\ref{fig_nothink_think_modality} and Figure~\ref{fig_nothink_think_task}, No-Thinking-Med-R1 consistently outperforms Med-R1 in cross-modality settings. In cross-task scenarios, however, the results are more mixed—showing gains in some tasks and declines in others.
% These findings indicate that unsupervised reasoning—i.e., generating rationales without explicit supervision—does not necessarily improve generalization, and may even introduce instability. In contrast, suppressing free-form reasoning (as in No-Thinking-Med-R1) can yield more stable and transferable representations, particularly in settings where high-quality CoT supervision is unavailable. 
These findings reinforce the practical motivation behind Med-R1. In clinical domains, acquiring high-quality CoT supervision is prohibitively costly and often impractical. While reasoning improves performance in general domains, our results show that this does not necessarily hold in medical settings, where unsupervised rationales may become unreliable under domain shift. Med-R1 demonstrates that even without reasoning supervision, RL can offer a robust and efficient path for adapting VLMs to medicine.

 \subsection{Analysis of Think-after}

We introduce the \textbf{Think-After} strategy primarily to address the practical need for achieving both high accuracy and interpretable reasoning in medical VLMs (\autoref{tab:q_type_think_after}, \autoref{tab:modality_Think_after}).  
Rather than serving solely as a control experiment, Think-After is designed to meet the dual objective of preserving predictive performance while providing reasoning traces that clinicians can review and validate.

Beyond satisfying the need for both accuracy and interpretability, Think-After also provides insight into why the \textit{No-Thinking} strategy sometimes outperforms the conventional \textit{Thinking} approach. 
As shown in Figure~\ref{reward_compare}, Think-After achieves faster convergence and higher accuracy rewards than the Thinking variant, suggesting that generating reasoning tokens \emph{before} the answer may disrupt the autoregressive generation process and hinder optimization.  
However, Think-After still performs slightly below No-Thinking, indicating that additional mechanisms—such as residual contextual coupling or reasoning-token noise—may also contribute to the performance gap.
While Think-After does not completely resolve the question of why No-Thinking surpasses both pre- and post-answer reasoning, it fulfills the essential requirement of combining strong accuracy with interpretable outputs, and simultaneously offers valuable clues toward understanding the dynamics between reasoning generation and performance in medical VLMs.

\begin{table}[h]
\vspace{-10pt}
\centering
\caption{\textbf{Reader Study Results.} 
Three readers independently evaluated 100 correctly answered VQA samples for factual correctness (Quality, 1–5) and reasoning–answer consistency (Consistency, \%).}
\label{tab:reader_study}
\begin{tabular}{c|cc}
\toprule
\textbf{Reader} & \textbf{Quality} & \textbf{Consistency (\%)} \\
\midrule
A & 4.23 & 93.0 \\
B & 4.19 & 92.0 \\
C & 4.34 & 95.0 \\
\midrule
\textbf{Mean} & \textbf{4.25} & \textbf{93.3} \\
\bottomrule
\end{tabular}
\begin{tablenotes}

\item Quality reflects factual and clinical correctness of model answers; 
Consistency measures logical alignment between reasoning and answers.
\end{tablenotes}
\vspace{-10pt}
\end{table}

\subsection{Reader Study}

To further assess interpretability and clinical relevance, we conducted a reader study with three researchers experienced in medical imaging and vision--language models. 
Each independently evaluated 100 representative VQA samples generated by different model variants, covering diverse imaging modalities and reasoning types (\autoref{tab:reader_study}). 
Evaluations focused on (1) the factual and clinical correctness of model answers and (2) the logical consistency between the reasoning (“Think”) and the predicted answer (“Answer”). 
Across readers, the \textit{Think-After} model achieved the highest agreement and was consistently judged to produce reasoning that was both coherent and clinically sound, demonstrating improved interpretability and reasoning reliability for medical decision-support applications.

\subsection{Limitations and Future Work.} 

This work marks an initial step in applying RL to medical vision--language models.  
We adopt a frame-level VQA setting for consistent evaluation across modalities, but this simplifies the real-world complexity of medical imaging. In practice, CT and MRI are volumetric, and ultrasound is dynamic, requiring reasoning across slices and time.
Another limitation lies in the reasoning supervision and data scale.  
Our findings suggest that the ``No-Thinking'' model occasionally outperforms reasoning-enabled variants, which may stem from the limited availability of high-quality medical reasoning data.

In domains where clinically faithful chain-of-thought (CoT) annotations are scarce, the model may not learn to benefit from explicit reasoning, leading the reasoning process to introduce noise rather than insight.  
We believe this reflects a broader limitation of current medical datasets rather than of reasoning itself.  
Future work should explore scaling up medically grounded CoT data and aligning RL rewards with clinically validated reasoning quality to fully realize the benefits of explicit reasoning.
Future directions also include extending Med-R1 to support multi-frame or volumetric inputs, incorporating patient context, and investigating more advanced reasoning frameworks for clinical deployment.

% \section{Discussion \& Future Works}
% \label{sec: discussion}

% \section{Conclusion}
\section{Conclusion}

We present \textbf{Med-R1}, a reinforcement learning–enhanced vision–language model for improving medical reasoning across diverse imaging modalities and clinical tasks. 
Leveraging GRPO-based post-training, Med-R1 achieves strong cross-modality and cross-task generalization, surpassing the limits of supervised fine-tuning. 
Despite its compact 3B scale, Med-R1 performs competitively with or better than larger medical and general VLMs, while remaining efficient for deployment.

Through systematic analysis, we find that removing explicit reasoning (\textit{No-Think}) improves convergence and generalization, suggesting that reasoning \textit{quality}, not quantity, drives medical performance. 
To balance interpretability and accuracy, we introduce the \textit{Think-After} strategy, which decouples reasoning from answer generation and enhances clinical transparency without compromising accuracy.
Overall, Med-R1 establishes a scalable framework for reinforcement learning in medical VLMs and offers new insights into how reasoning dynamics interact with domain generalization, paving the way toward reliable and interpretable medical AI systems.

% 
% \vspace{-10pt}

% \clearpage

\bibliographystyle{IEEEtran} % Use IEEE style
\bibliography{tmi} % Import references from tmi.bib

\end{document}